\documentclass[10pt,twocolumn,letterpaper]{article}

\usepackage[pagenumbers]{cvpr} 
\usepackage[accsupp]{axessibility}  

\usepackage{graphicx}
\usepackage{amsmath}
\usepackage{amssymb}
\usepackage{booktabs}

\usepackage[pagebackref,breaklinks,colorlinks]{hyperref}

\usepackage[capitalize]{cleveref}
\crefname{section}{Sec.}{Secs.}
\Crefname{section}{Section}{Sections}
\Crefname{table}{Table}{Tables}
\crefname{table}{Tab.}{Tabs.}

\def\cvprPaperID{8115} 
\def\confName{CVPR}
\def\confYear{2022}

\usepackage{times}
\usepackage{epsfig}
\usepackage{graphicx}
\usepackage{amsmath}
\usepackage{amssymb}
\usepackage{enumitem}

\usepackage{booktabs}
\usepackage{multirow}
\usepackage[table]{xcolor}
\usepackage{float}
\definecolor{Gray}{gray}{0.93}
\usepackage{pifont}

\newcommand{\nocaps}{\texttt{nocaps}}

\newcommand{\vv}{{\boldsymbol v}}
\newcommand{\tv}{{\boldsymbol t}}
\newcommand{\wv}{{\boldsymbol w}}
\newcommand{\xv}{{\boldsymbol x}}

\newcommand{\Vmat}{{\bf V}}
\newcommand{\Tmat}{{\bf T}}
\newcommand{\Wmat}{{\bf W}}
\newcommand{\Rmat}{{\bf R}}
\newcommand{\Ymat}{{\bf Y}}
\newcommand{\Lcal}{{\mathcal{L}}}

\newcounter{magicrownumbers}
\newcommand\rownumber{\stepcounter{magicrownumbers}\arabic{magicrownumbers}}

\begin{document}

\title{Scaling Up Vision-Language Pre-training for Image Captioning}

\author{Xiaowei Hu, Zhe Gan, Jianfeng Wang, Zhengyuan Yang, \\
Zicheng Liu, Yumao Lu, Lijuan Wang\\
Microsoft \\
{\tt\small \{xiaowei.hu,zhe.gan,jianfw,zhengyang,zliu,yumaolu,lijuanw\}@microsoft.com}
}

\maketitle

\begin{abstract}
    In recent years, we have witnessed significant performance boost in the image captioning task based on vision-language pre-training (VLP).
   Scale is believed to be an important factor for this advance. However, most existing work only focuses on pre-training transformers with moderate sizes (e.g., 12 or 24 layers) on roughly 4 million images.
   In this paper, we present LEMON~\includegraphics[height=11pt]{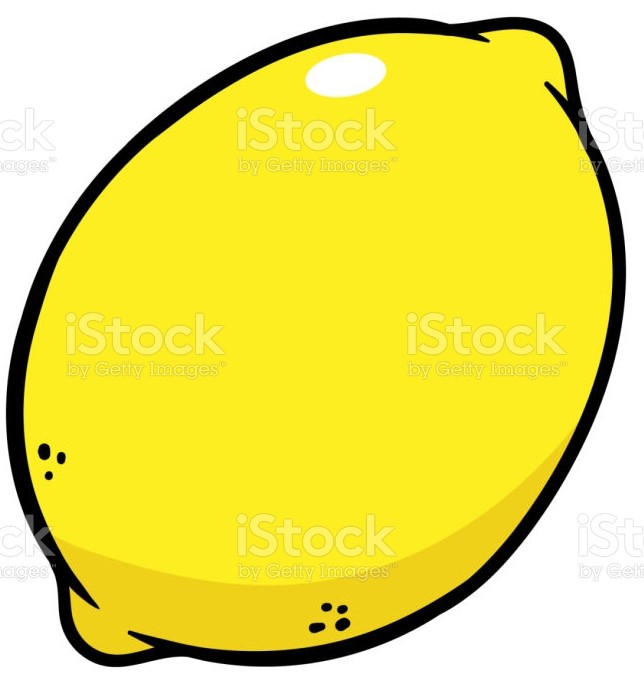}, a LargE-scale iMage captiONer, and provide the first empirical study on the scaling behavior of VLP for image captioning. We use the state-of-the-art VinVL model as our reference model, which consists of an image feature extractor and a transformer model,  
   and scale the transformer both up and down, with model sizes ranging from 13 to 675 million parameters.  
   In terms of data, we conduct experiments with up to 200 million image-text pairs which are automatically collected from web based on the alt attribute of the image (dubbed as ALT200M
   ).
   Extensive analysis helps to characterize the performance trend as the model size and the pre-training data size increase. We also compare different training recipes, especially for training on large-scale noisy data. As a result, LEMON achieves new state of the arts on several major image captioning benchmarks, including COCO Caption, \nocaps,~and Conceptual Captions. 
   We also show LEMON can generate captions with long-tail visual concepts when used in a zero-shot manner.
\end{abstract}


\vspace{-2mm}
\section{Introduction}

\begin{figure}[t]
\begin{center}
\includegraphics[trim=5 15 5 0, clip,width=0.49\textwidth]{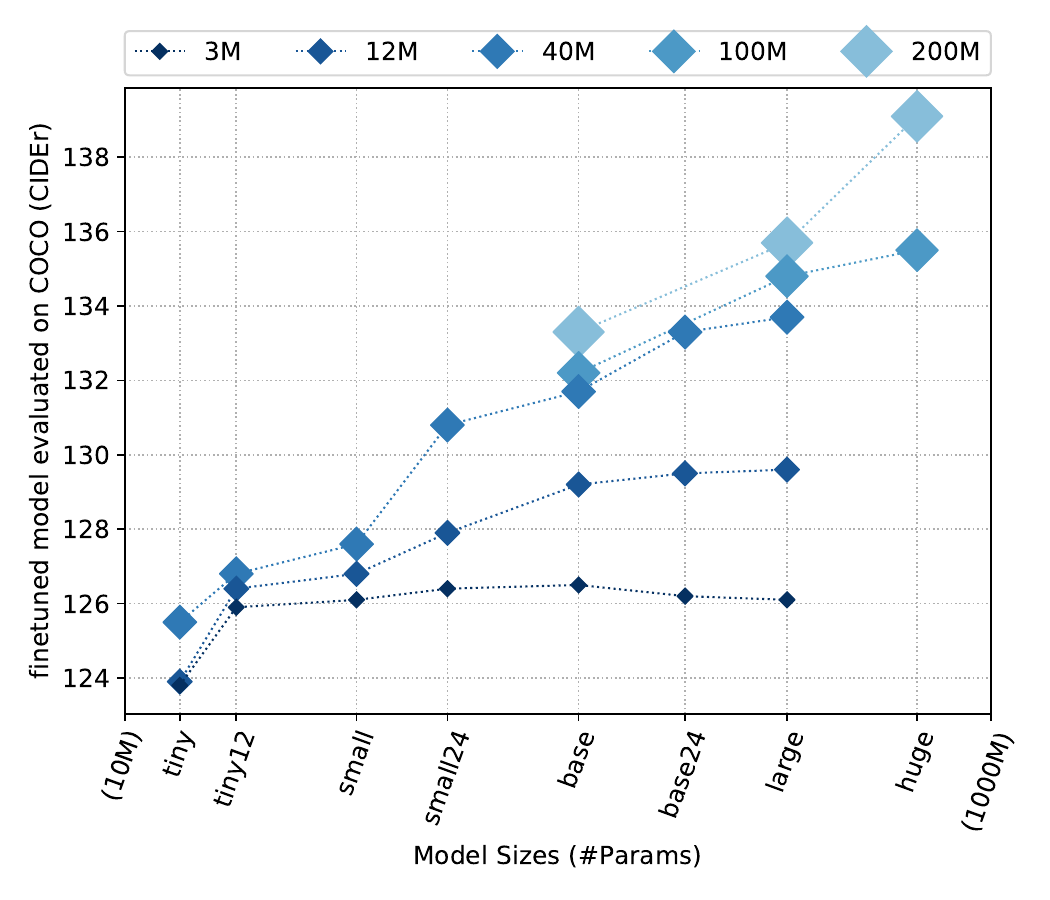}
\vspace{-4mm}
\caption{\textbf{Image captioning performance on COCO when upscaling model for each dataset size}. The x-axis plots the number of parameters for each model size (\emph{e.g.}, tiny, small, huge) in a logarithmic scale. The definition of model sizes is detailed in Table \ref{tab:arch_details}. Increasing the model size is not significantly beneficial at small pre-training dataset scales. However, when we use sufficiently large datasets, we see strong performance boost from a larger model. }
\label{fig:model_size}
\end{center}
\vspace{-6mm}
\end{figure}


\begin{figure*}[t!]
    \centering
    \subfloat[\centering finetuned and evaluated on COCO]{{\includegraphics[trim=15 0 55 0,clip,height=0.27\textheight]{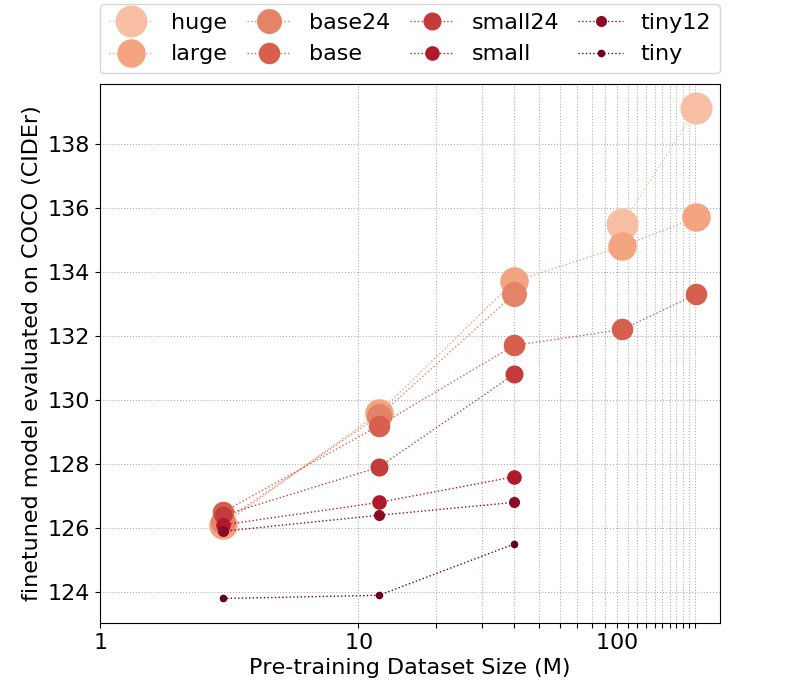} }}%
    \qquad
    \subfloat[\centering finetuned on COCO, evaluated on nocaps]{{\includegraphics[trim=85 3 110 0,clip,height=0.27\textheight]{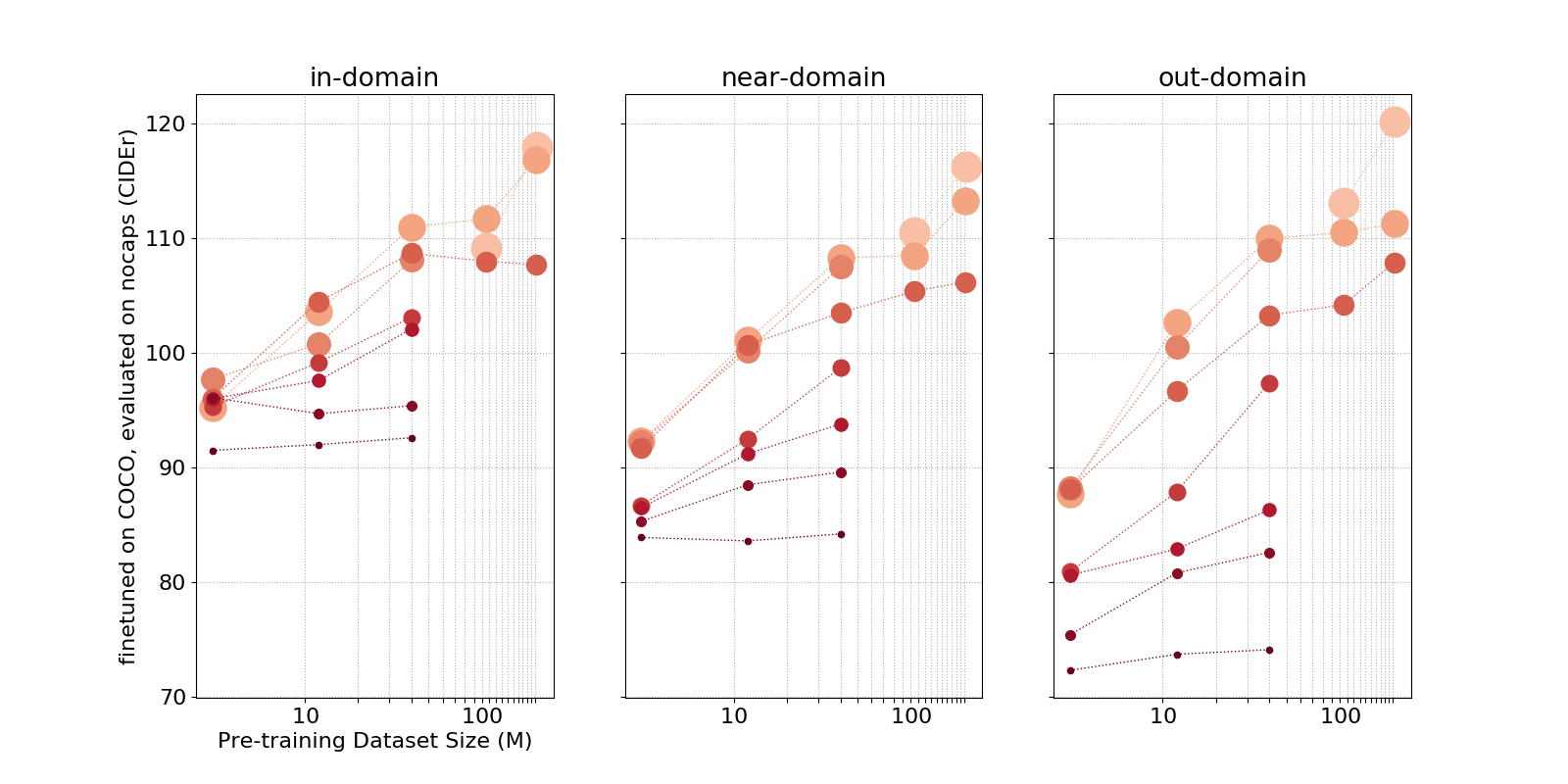} }}%
\caption{\textbf{Image captioning performance in data upscaling for each model size}. The x-axis shows the number of image-text pairs used in pre-training. The y-axis shows the evaluation score (CIDEr) on COCO ``Karpathy'' test split and \nocaps~validation set, respectively. The models are first pre-trained, then finetuned on COCO caption training split. Note that x-axis is plotted in a logarithmic scale.}%
\label{fig:data_size}%
\vspace{-2mm}
\end{figure*}

Recent advances in image captioning~\cite{chen2015microsoft,sharma2018conceptual,agrawal2019nocaps}
can be largely attributed to vision-language pre-training (VLP)~\cite{tan2019lxmert,lu2019vilbert,su2019vl,li2019visualbert}, the current prevailing training paradigm for vision-language (VL) research.
VLP~\cite{chen2019uniter} is usually conducted on a combined image-text dataset comprising of several or tens of millions images in total, \eg, Visual Genome~\cite{krishna2017visual}, SBU~\cite{ordonez2011im2text} and Conceptual Captions~\cite{sharma2018conceptual,changpinyo2021conceptual}.
While previous studies~\cite{zhou2020unified,li2020oscar,zhang2021vinvl} have analyzed various choices of pre-training objectives and model architectures, it remains unclear to what extent the pre-training dataset would impact the performance, and how it correlates with different model settings. Along the journey of pushing the limit of VLP, it becomes increasingly important to answer this question.

\emph{Scale} is believed to be an important ingredient in attaining excellent performance~\cite{radford2021learning,jia2021scaling,wang2021simvlm}.
Recent work has investigated the Pareto frontier of training transformer models, often referred to as the neural scaling law, in the domains of natural language processing~\cite{kaplan2020scaling,brown2020language,tay2021scale} and computer vision~\cite{henighan2020scaling,zhai2021scaling}, via unsupervised or weakly-supervised learning methods. These studies have observed consistent benefits of increasing the model size to billions of parameters, given billion magnitude of pre-training data available.

More recently, contrastive image-text pre-training~\cite{radford2021learning,jia2021scaling} has also been scaled up to 400 million and 1.8 billion data sizes
for image representation learning and image-text retrieval. 
Both CLIP~\cite{radford2021learning} and ALIGN~\cite{jia2021scaling} employ two individual networks to encode the image and the text separately for alignment, which well fits the image-text retrieval task, but
little is known about the scaling properties when it comes to image captioning.

\begin{table*}[t!]
\small
\begin{center}
\begin{tabular}{l@{\hspace{2mm}}|c@{\hspace{2mm}}c|c@{\hspace{2mm}}c@{\hspace{2mm}}|l@{\hspace{2mm}}c}
\toprule
\multirow{2}{*}{Dataset} & \multirow{2}{*}{\#images (M)} & \multirow{2}{*}{\#cap./image} & \multicolumn{2}{c|}{Unigram} & \multicolumn{2}{c}{Caption lengths} \\ 
& & & \#unique &  \#unique in $0.1\%$ tail  & mean $\pm$ std & P5\%/50\%/95\%\\ \midrule
COCO Caption~\cite{chen2015microsoft} & $0.1$ & $5$ &  $19,264$ &  $1,184$ & $10.44 \pm 2.24$ & $8/10/14$ \\
CC3M~\cite{sharma2018conceptual} & $3.1$ & $1$&  $49,638$ &  $22,677$ & $10.25 \pm 4.64$ & $5/9/19$ \\
CC12M~\cite{changpinyo2021conceptual}& $12.2$& $1$ &  $1,319,284$  & $193,368$ & $17.17 \pm 12.76$ & $6/13/43$ \\
ALT200M (Ours) & $203.4$& $1$ &  $2,067,401$ & $1,167,304$ & $13.01 \pm 8.85$ & $2/11/27$ \\
\bottomrule
\end{tabular}
\end{center}
\vspace{-5mm}
\caption{\textbf{Statistics of existing and our collected datasets}. The number of images in CC3M and CC12M are calculated for valid RGB images at the time we downloaded them. The unigrams are counted and sorted by occurrences from large to small to form a distribution curve for each dataset. Our dataset features much more long-tail concepts, as indicated by the number of unigrams included in the $0.1\%$ distribution tail. The datasets used in CLIP~\cite{radford2021learning} and ALIGN~\cite{jia2021scaling} are not included, since we do not know the corresponding statistics.}
\label{tab:data_stats}
\vspace{-3mm}
\end{table*}

To study the characteristics of this scaling trend on the captioning task, 
we first construct a large-scale image-text dataset (dubbled as ALT200M), 
consisting of up to 200 million image-text pairs
from web based on the alt attribute of the images.
Then, we conduct extensive experiments to scale VLP for image captioning from both the \emph{data} and \emph{model} perspectives, and name our model as LEMON~\includegraphics[height=10pt]{figs/lemon.jpg}, short for a LargE-scale 
iMage captiONer.
To simulate the process of data scaling, we create multiple subsets of ALT200M, ranging from $3$ to $200$ million. 
In terms of model, we use the state-of-the-art image captioning model VinVL~\cite{zhang2021vinvl} as our reference model, composed of an image feature extractor and a transformer model. 
We adapt the pre-training task to be consistent with the captioning task,
and then scale the width and depth of the transformer model with the number of parameters ranging from $13$ (\emph{i.e.}, tiny) to $675$ (\emph{i.e.}, huge) millions. 
Combining different models and pre-training data sizes, we summarize our results in \Cref{fig:model_size} and \ref{fig:data_size}, which characterize the linear-logarithmic scaling trend. Larger models tend to benefit more when we have more than $10$ million data for pre-training. However, with only $3$ million data, the performance starts to saturate early as the model size increases.  
Moreover, we also investigate other design choices of VLP, \eg, model architectures and training objectives. 
Our contributions are summarized as follows.
\begin{itemize}[leftmargin=*]
    \vspace{-2mm}
    \item We present the VLP scaling rule for image captioning. Not only does this prove the effectiveness of learning from large-scale noisy data, but it also sheds lights on how performance can be efficiently improved by increasing the model and pre-training data sizes together to avoid a saturation plateau.
    \vspace{-2mm}
    \item We achieve new state-of-the-art results for image captioning across several major benchmarks, including COCO Caption, \nocaps,~and Conceptual Captions.
\end{itemize}

\section{Related Work}
\paragraph{Vision-Language Pre-training.}
Since the birth of ViLBERT~\cite{lu2019vilbert} and LXMERT~\cite{tan2019lxmert}, we have witnessed a boom of methods for vision-language pre-training~\cite{chen2019uniter,li2019visualbert,li2019unicoder,su2019vl,yu2020ernie,hu2020vivo,yang2020tap,cho2021unifying}. Prominent examples include UNITER~\cite{chen2019uniter}, VL-BERT~\cite{su2019vl}, OSCAR~\cite{li2020oscar}, UNIMO~\cite{li2020unimo}, and VinVL~\cite{zhang2021vinvl}. Along the journey of VLP, researchers have investigated different training strategies~\cite{gan2020large,lu201912}, robustness~\cite{li2020closer}, compression~\cite{wang2020minivlm,fang2021compressing,gan2021playing}, probing analysis~\cite{cao2020behind,li2020does}, and the extension to video-text modeling~\cite{sun2019videobert,sun2019contrastive,li2020hero,zhu2020actbert,lei2021less}. More recently, instead of using object detectors for image feature extraction, end-to-end VLP based on convolution networks and transformers are becoming popular~\cite{huang2020pixel,huang2021seeing,kim2021vilt,xue2021probing,li2021align}.

However, as another important factor in achieving superior performance, 
the scaling behavior of VLP is less studied. While most works pre-train transformer of base/large sizes on no more than $4$M images, we train models from tiny to huge, on up to $200$M images. 
CLIP~\cite{radford2021learning} and ALIGN~\cite{jia2021scaling} scaled up contrastive pre-training to 400M and 1.8B images, and SimVLM~\cite{wang2021simvlm} further use 1.8B images for prefix language modeling pre-training. However, CLIP and ALIGN focus on image-text retrieval, while SimVLM did not study its scaling behavior w.r.t. pre-training data sizes. Compared with them, we focus on image captioning, provide a more comprehensive study on the scaling behavior via altering data and model sizes, and show that by using 200M images, we can outperform SimVLM on image captioning. 

\vspace{-3mm}
\paragraph{Scaling Law.}
With the success of large-scale pre-trained models in both  the language and vision domains, there has been a surging research interest in discovering the empirical scaling law of these models. \cite{kaplan2020scaling} presented that the language model performance scales as power-law across many orders of magnitude with dataset size, model size, and computation used in training. \cite{henighan2020scaling} further studied the scaling of autoregressive generative modeling. Aside from the model size, \cite{tay2021scale} showed that the model shape also matters for efficient transfer from upstream pre-training to downstream finetuning. In the vision domain, \cite{zhai2021scaling} scaled a series of vision transformer models evaluated on image classification tasks. While the scaling protocols have been investigated for many NLP and vision tasks, we are the first to study the scaling behavior of VLP for image captioning, and push multimodal transformer pre-training to a much larger scale.

In Appendix, we also provide a detailed related work review on non-pretraining-based image captioning methods.

\section{Method}

In this section, we present the pre-training dataset in Section~\ref{sec:dataset}, the model structure in Section~\ref{sec:model}, and training objective in Section~\ref{sec:training}.

\subsection{Pre-training Dataset} \label{sec:dataset}

We construct a data collection pipeline to crawl the images from the Internet and the associated \textit{alt} attribute, which usually provides the description of the image content.
In order to scale up easily, we follow the natural distribution of images 
without re-balancing, and apply only minimal rule-based filtering.
We keep images with the longer side more than $200$ pixels and aspect ratio smaller than $3$. As some alt-texts are too long, we split them up by punctuation marks, such as period and exclamation mark, and select the longest part. To filter out some rare or misspelled words, we build a vocabulary of unigrams with English Wikipedia titles and body text. We remove unigrams that are present less than $5$ times, resulting in approximately $250$ million unique unigrams. We remove the alt-text if any of its unigrams cannot be found in the vocabulary. Afterwards,
we count the frequency of all the remaining sentences, and filter out some boilerplate sentences that are too generic, \textit{e.g.}, stock image, 3D illustration, vector photo. For the sake of privacy, we use a Named Entity Recognition model spaCy\footnote{\url{https://github.com/explosion/spaCy}} to identify person and location names, and replace them with special tokens \textlangle PERSON\textrangle, \textlangle LOC\textrangle, respectively.
At last, we perform duplication check on all the collected images to ensure that they do not overlap with existing test sets, such as COCO, \nocaps, and Conceptual Captions.

\begin{figure}[t]
\begin{center}
\includegraphics[trim=0 0 0 0, clip,width=0.45\textwidth]{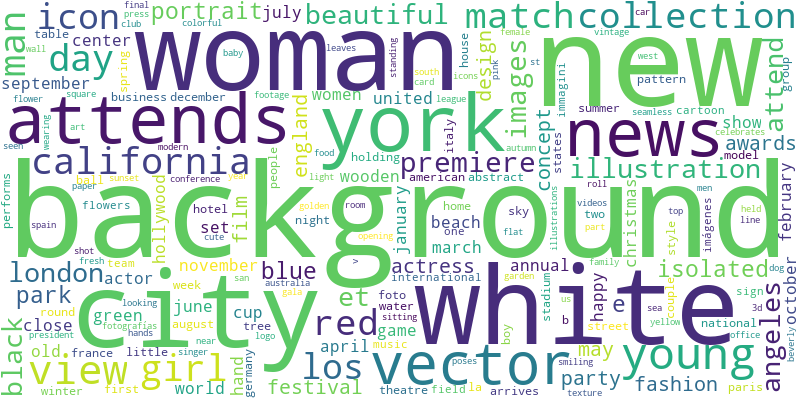}
\vspace{-2mm}
\caption{\textbf{Word cloud of the top 200 words} in our pre-training dataset ALT200M, excluding the stop words, \eg, a, the, of, \etc. }
\label{fig:word_cloud}
\end{center}
\vspace{-5mm}
\end{figure}

The final dataset, named as ALT200M, contains more than $200$ million images, each corresponding to one alt-text. The word cloud of $200$ most frequent words is visualized in \Cref{fig:word_cloud}.
As shown in \Cref{tab:data_stats}, compared to CC12M, ALT200M has nearly $16\times$ more images. 
The vocabulary is almost doubled. We observe that $56\%$ of unigrams sum up to only $0.1\%$ of total occurrences, characterizing an extremely long tail of rarely occurring unigrams.
The average length of the captions is $13.01$, more than
that of the COCO caption dataset ($10.44$).
We also observe that our dataset contains much more shorter captions with only $2$ or $3$ unigrams. This indicates a shift in the distribution of captions from pre-training to finetuning.

Besides CC12M, there also exist some other large-scale image-text datasets, such as WIT~\cite{srinivasan2021wit}, WenLan~\cite{huo2021wenlan}, LAION-400M~\cite{schuhmann2021laion}, and the datasets used in CLIP~\cite{radford2021learning} and ALIGN~\cite{jia2021scaling}.
More detailed discussions on them are provided in Appendix.


\begin{table}
\small
\centering
\begin{tabular}{ccccccc}
\toprule
\rotatebox{90}{Model} & \rotatebox{90}{Layers} & \rotatebox{90}{Width} & \rotatebox{90}{MLP} & \rotatebox{90}{Heads} & \rotatebox{90}{Param (M)} & \rotatebox{90}{FLOPs}\\
\midrule
tiny & $6$ & $256$ & $1024$ & $4$ & $13.4$ & $1.1$\\
tiny12 & $12$ & $256$ & $1024$ & $4$ & $18.1$ & $1.5$\\
small & $12$ & $384$ & $1536$ & $6$ & $34.3$ & $2.9$\\
small24 & $24$ & $384$ & $1536$ & $6$ & $55.6$ & $4.8$\\
base & $12$ & $768$ & $3072$ & $12$ & $111.7$ & $9.5$\\
base24 & $24$ & $768$ & $3072$ & $12$ & $196.7$ & $16.8$\\
large & $24$ & $1024$ & $4096$ & $16$ & $338.3$ & $28.9$\\
huge & $32$ & $1280$ & $5120$ & $16$ & $675.4$ & $57.7$\\
\bottomrule
\end{tabular}
\vspace{-2mm}
\caption{\textbf{Details of model architecture}. FLOPs are calculated via taking $50$ image region features and $35$ text tokens as input in one forward pass. The dimension of image region feature is $2054$, which is mapped to the transformer width via a linear layer.}
\vspace{-3mm}
\label{tab:arch_details}
\end{table}

\subsection{VLP Model for Captioning} \label{sec:model}
We use the pre-trained Faster R-CNN detector from~\cite{zhang2021vinvl} to extract image region features, which are concatenated with scaled bounding boxes as position encoding. Following~\cite{li2020oscar,zhang2021vinvl}, we also add the detected object tags as input. The text input, including the caption and objects tags, are tokenized by WordPiece, with a vocabulary of $30522$ tokens. A multi-layer transformer model is used for multimodal fusion, which consists of a stack of encoder layers, each of which has a multi-head self-attention (MSA) layer followed by a feed-forward layer.
To enable text generation with the encoder layers, we use the sequence-to-sequence attention mask~\cite{zhou2020unified} in each self-attention layer for the captioning module.
Specifically, the input consists of image embeddings $\Vmat = \{\vv_i\}_{i=1}^{N}$, object tag embeddings $\Tmat = \{\tv_j\}_{j=1}^{M}$, and token embeddings for the caption
$\Wmat = \{\wv_k\}_{k=1}^{L}$, where $N, M, L$ are the number of image regions, tags, and caption tokens, respectively. The corresponding outputs are:
\begin{align}
    \Rmat_{v_i} &:= \text{MSA}(\vv_i, \Vmat \cup \Tmat)\,, \\
    \Rmat_{t_j} &:= \text{MSA}(\tv_j, \Vmat \cup \Tmat)\,, \\
    \Rmat_{w_k} &:= \text{MSA}(\wv_k, \Vmat \cup \Tmat \cup \{\wv_l\}_{l=1}^{k})\,,
\end{align}
where $\text{MSA}(\xv, \Ymat)$ is the MSA layer with $\xv$ mapped to query, and $\Ymat$ mapped to key/value. $\cup$ means concatenation of matrices, and the index of $\Rmat_{v_i}$ denotes the position corresponding to $\vv_i$. The output representation is fed into the next layer, or used for prediction at the end.
In this way, during inference, the model can decode the token from left to right in an auto-regressive manner.
To study the scaling trend, we experiment with $8$ model configurations, ranging from ``tiny'' of $13$M parameters to ``huge'' of $674$M parameters, detailed in \Cref{tab:arch_details}. 

\begin{figure}[t]
\begin{center}
\includegraphics[trim=0 140 480 0, clip,width=0.4\textwidth]{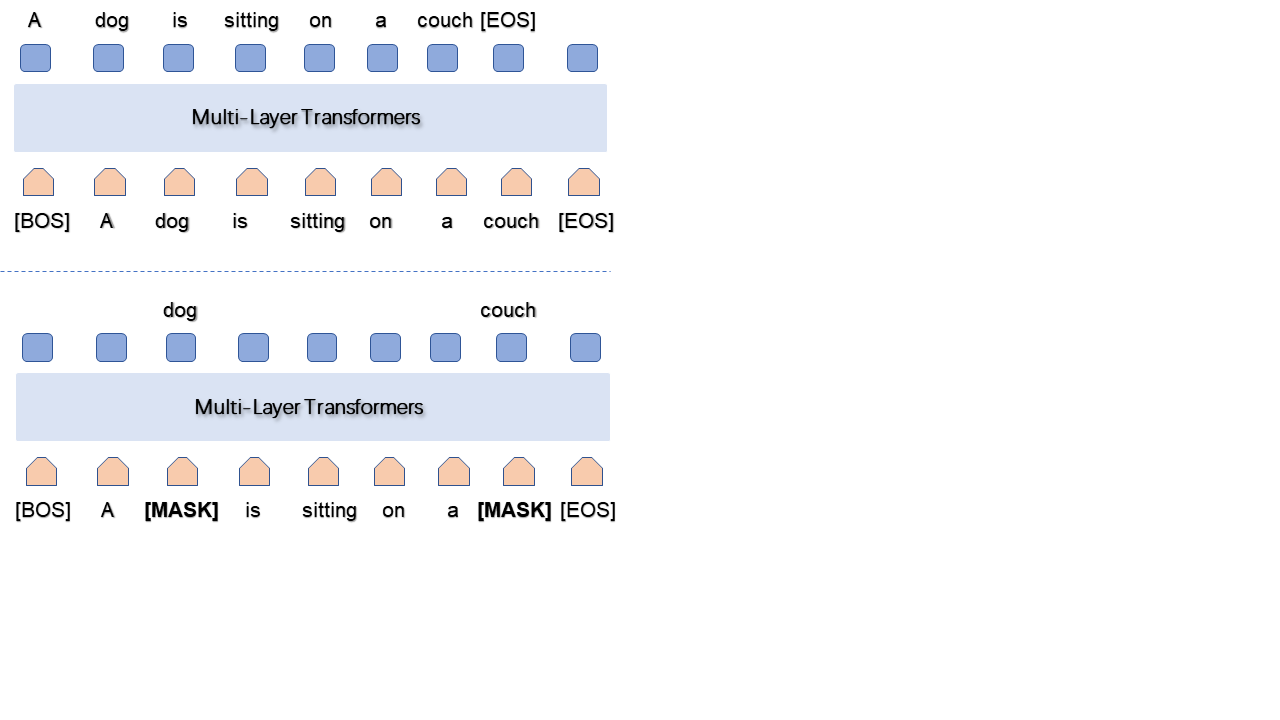}
\vspace{-2mm}
\caption{\textbf{Comparison of training objectives}. (Top) Language Modeling (LM), to predict the next token at each position. (Bottom) Masked Language Modeling (MLM), to predict the masked and/or possibly polluted tokens at the masked positions. Both use causal masking for model training.}
\label{fig:lm}
\end{center}
\vspace{-5mm}
\end{figure}

\begin{table*}[ht]
\small
\begin{center}
\begin{tabular}{l@{\hspace{4mm}} l@{\hspace{2mm}}|l@{\hspace{2mm}}|c@{\hspace{2mm}}c@{\hspace{2mm}}|c@{\hspace{2mm}}c@{\hspace{2mm}}|c@{\hspace{2mm}}c@{\hspace{2mm}}|c@{\hspace{2mm}}c}
\toprule
\multirow{2}{*}{\# } & \multirow{2}{*}{Model } & \multirow{2}{*}{Pre-training data } & \multicolumn{2}{c|}{in-domain} & \multicolumn{2}{c|}{near-domain} & \multicolumn{2}{c|}{out-of-domain} & \multicolumn{2}{c}{overall}\\ 
& & & CIDEr & SPICE & CIDEr & SPICE & CIDEr & SPICE & CIDEr & SPICE \\ \midrule
\multicolumn{10}{c}{Validation Set} \\ \hline
\rownumber & Encoder-Decoder~\cite{changpinyo2021conceptual} & CC3M~\cite{sharma2018conceptual}  & $81.8$ & $11.6$ & $73.7$ & $11.1$ & $65.3$ & $10.1$ & $73.2$ & $11.0$ \\ 
\rownumber &  & CC12M~\cite{changpinyo2021conceptual} & $88.3$ & $12.3$ & $86.0$ & $11.8$ & $91.3$ & $11.2$ & $87.4$ & $11.8$ \\ 
\rownumber &  & CC3M+CC12M & $92.6$ & $12.5$ & $88.3$ & $12.1$ & $94.5$ & $11.9$ & $90.2$ & $12.1$ \\ 
\midrule
\rownumber & VinVL$_{base}$$^{*}$~\cite{zhang2021vinvl} & N/A & $96.8$ & $13.5$ & $90.7$ & $13.1$ & $87.4$ & $11.6$ & $90.9$ & $12.8$ \\
\rownumber & VinVL$_{base}$$^{\dagger}$ & $5.65$M combined & $103.1$ & $14.2$ & $96.1$ & $13.8$ & $88.3$ & $12.1$ & $95.5$ & $13.5$ \\
\rownumber & VinVL$_{large}$$^{\dagger}$ & $5.65$M combined & $106.3$ & $14.5$ & $98.0$ & $14.0$ & $88.8$ & $12.6$ & $97.3$ & $13.8$ \\
\midrule
\rownumber & SimVLM$_{huge}$~\cite{wang2021simvlm} & $1.8$B & $113.7$ &  - & $110.9$ & - & $115.2$ & - & $112.2$ & - \\
\midrule
\rownumber & LEMON$_{base}$ & N/A & $91.4$ & $13.3$ & $81.4$ & $12.5$ & $62.6$ & $10.6$ & $79.0$ & $12.3$ \\
\rownumber & LEMON$_{base}$ & CC3M & $96.0$ & $13.8$ & $91.7$ & $13.2$ & $88.1$ & $11.8$ & $91.6$ & $13.0$ \\
\rownumber & LEMON$_{base}$ & CC12M & $104.5$ & $14.6$ & $100.7$ & $14.0$ & $96.7$ & $12.4$ & $100.4$ & $13.8$ \\
\rownumber & LEMON$_{large}$ & CC12M & $103.6$ & $14.4$ & $101.1$ & $13.8$ & $102.7$ & $12.6$ & $101.8$ & $13.6$ \\
\rowcolor{Gray}
\rownumber & LEMON$_{base}$ & ALT200M & $107.7$ & $14.7$ & $106.2$ & $14.3$ & $107.9$ & $13.1$ & $106.8$ & $14.1$ \\
\rowcolor{Gray} 
\rownumber & LEMON$_{large}$ & ALT200M & $116.9$ & $\bf 15.8$ & $113.3$ & $15.1$ & $111.3$ & $14.0$ & $113.4$ & $15.0$ \\
\rowcolor{Gray} 
\rownumber & LEMON$_{huge}$ & ALT200M & $\bf 118.0$ & $15.4$ & $\bf 116.3$ & $\bf 15.1$ & $\bf 120.2$ & $\bf 14.5$ & $\bf 117.3$ & $\bf 15.0$  \\
\bottomrule
\multicolumn{9}{c}{Test Set} \\ \hline
\rownumber & Human & & $80.6$ & $15.0$ & $84.6$ & $14.7$ & $91.6$ & $14.2$ & $85.3$ & $14.6$ \\
\midrule
\rownumber & SimVLM$_{base}$ & $1.8$B & - & - & - & - & - & - & $94.8$ & $13.1$\\
\rownumber & SimVLM$_{large}$ & $1.8$B & - & - & - & - & - & - & $108.5$ & $14.2$\\
\rownumber & SimVLM$_{huge}$$^{\ddagger}$ & $1.8$B & $109.0$ & $14.6$ & $110.8$ & $14.6$ & $109.5$ & $\bf 13.9$ & $110.3$ & $14.5$\\
\midrule
\rowcolor{Gray} 
\rownumber & LEMON$_{large}$ & ALT200M & $111.2$ & $\bf 15.6$ & $112.3$ & $\bf 15.2$ & $105.0$ & $13.6$ & $110.9$ & $\bf 15.0$ \\
\rowcolor{Gray} 
\rownumber & LEMON$_{huge}$ & ALT200M & $\bf 112.8$ & $15.2$ & $\bf 115.5$ & $15.1$ & $\bf 110.1$ & $13.7$ & $\bf 114.3$ & $14.9$ \\
\bottomrule
\end{tabular}
\end{center}
\vspace{-4mm}
\caption{\textbf{Results on \nocaps~validation and test sets}. All our models are trained with cross-entropy loss only, without CIDEr optimization. The VinVL model with $*$ is not pre-trained, but use SCST+CBS as reported in the paper. The VinVL results with $\dagger$ are reproduced by us via finetuning from the released checkpoints, which are pre-trained on the combined datasets including $5.65$M images, $2.5$M QAs, $4.68$M captions and $1.67$M pseudo-captions. The numbers with $\ddagger$ are copied from the \nocaps~leaderboard.}
\vspace{-3mm}
\label{tab:nocaps_caption}
\end{table*}

\subsection{Training Objective} \label{sec:training}
\label{sec:training_obj}
While bidirectional Masked Language Modeling (MLM) has been widely used in both language and vision-language pre-training, its bidirectional nature makes it sub-optimal for text generation.
In contrast to VLP works that are mostly evaluated on VL understanding tasks, we use sequence-to-sequence MLM for generation tasks.
During training, we randomly mask out $15\%$ of caption tokens following BERT~\cite{devlin2018bert} to form a ``corrupted'' caption $\widetilde{\Wmat} = \{\tilde{\wv}_k\}_{k=1}^{L}$, where $\tilde{\wv}_k$ is either equal to $\wv_k$, or replaced with \texttt{[MASK]} token or another token sampled from vocabulary. The training loss is defined as:
\begin{align}
    \Lcal(\Wmat, \Vmat, \Tmat) 
    &= \sum_{k \in D} \text{CE}( \wv_k, \Rmat_{\tilde{w}_k}) \\
    &= \sum_{k \in D} (- \log p(\wv_k | \Vmat, \Tmat, \{\tilde{\wv}_l\}_{l=1}^{k})) \,, \nonumber
\end{align}
where $\text{CE}(\cdot, \cdot)$ is the cross-entropy loss with softmax, $D$ is the subset of masked positions. The loss for the recovery of the possibly polluted tokens by intuition fits into the scenario of training with noisy captions. Note that we use the same loss in pre-training and finetuning. During inference, at step $s$, given the previous predicted tokens $\{ \hat{w}_k \}_{k=1}^{s-1}$, we set $\tilde{w}_s$ to \texttt{[MASK]}, and $\tilde{w}_k = \hat{w}_k$ for $k<s$. Therefore, the generation process simulates recovering the \texttt{[MASK]} token at the end in each step. Since the representations of caption tokens do not depend on the subsequent tokens, the intermediate representations of predicted tokens can be saved to avoid duplicate computation, thereby making the generation efficient.
We also experimented with other model structures and training objectives, such as predicting the next token with language modeling, 
as shown in Figure~\ref{fig:lm} and will be detailed later in \Cref{sec:exp_analysis}.

\section{Experiments}
In this section, we first present our experimental setup in Section~\ref{sec:exp_setup}, and then detail our results in Section~\ref{sec:cap_results}, followed by comprehensive analysis in Section~\ref{sec:exp_analysis}. 

\subsection{Setup} \label{sec:exp_setup}

\paragraph{Datasets.} To measure the progress brought about by large-scale pre-training, we aim to evaluate the model's capability of describing varieties of (long-tail) visual concepts, which is essential for captioning in the wild. For this purpose, we choose \nocaps~\cite{agrawal2019nocaps} as the evaluation benchmark, which is developed to evaluate object captioning at scale. The dataset consists of $15100$ images from Open Images, and covers more than $600$ object categories, of which nearly $400$ of them are unseen from the training set in COCO~\cite{chen2015microsoft}.
Based on whether the image contains novel objects unseen in the COCO training set, the \nocaps~images are divided into three domains: ``in'', ``near'', and ``out''. None of the objects in the out-domain are seen in COCO.
This discrepancy raises the 
importance
of learning from external resources for recognizing novel objects, rather than relying on the clean and fully annotated captioning training data.
As the external training resources may vary for different methods,
in \Cref{tab:nocaps_caption}, we only compare our model with other models that also use extra image-caption pairs, and take the pre-training dataset size into account.
\begin{table*}
\small
\centering
\begin{tabular}{l|c|cccc|cccc}
\toprule
\multirow{2}{*}{Model } & \multirow{2}{*}{Pre-training data }  & \multicolumn{4}{c|}{Cross-entropy optimization} & \multicolumn{4}{c}{CIDEr optimization}\\
 &  & B@4 & M & C & S &  B@4 & M & C & S \\
\midrule
Encoder-Decoder~\cite{changpinyo2021conceptual} & CC12M & - & - & $110.9$ & - & - & - & - & - \\
\midrule
VinVL$_{base}$ & \multirow{2}{*}{$5.65$M combined} & $38.2$ & $30.3$ & $129.3$ & $23.6$ & $40.9$ & $30.9$ & $140.4$ & $25.1$ \\
VinVL$_{large}$ & & $38.5$ & $30.4$ & $130.8$ & $23.4$ & $41.0$ & $31.1$ & $140.9$ & $25.2$ \\
\midrule
SimVLM$_{base}$ &  & $39.0$ & $32.9$ & $134.8$ & $24.0$ &  - & - & - & -  \\
SimVLM$_{large}$ & 1.8B & $40.3$ & $33.4$ & $142.6$ & $24.7$ & - & - & - & -  \\
SimVLM$_{huge}$ &  & $40.6$ & $\bf 33.7$ & $143.3$ & $25.4$ & - & - & - & -  \\
\midrule
\rowcolor{Gray}
LEMON$_{base}$ &  & $40.3$ & $30.2$ & $133.3$ & $23.3$ & $41.6$ & $31.0$ & $142.7$ & $25.1$\\
\rowcolor{Gray}
LEMON$_{large}$ & ALT200M & $40.6$ & $30.4$ & $135.7$ & $23.5$ & $42.3$ & $31.2$ & $144.3$ & $25.3$\\
\rowcolor{Gray}
LEMON$_{huge}$ &  &  $41.5$ & $30.8$ & $139.1$ & $24.1$ & $\bf 42.6$ & $31.4$ & $\bf 145.5$ & $\bf 25.5$\\
\bottomrule
\end{tabular}
\vspace{-3mm}
\caption{\textbf{Results (single model) on COCO ``Karpathy'' test split}. B@4:
BLEU@4, M: METEOR, C: CIDEr, S: SPICE.}
\vspace{-4mm}
\label{tab:coco}
\end{table*}

\vspace{-3mm}
\paragraph{Implementation details.}
To study the scaling trend, we experiment with $8$ model configurations and $5$ pre-training data sizes. We train all the models from scratch if not otherwise specified.
In the pre-training, we do not include COCO or Visual Genome data, to exclude the possible impact of data quality when plotting the scaling trend, as these datasets are manually annotated instead of web collected.
To create pre-training dataset of different sizes, we randomly sample from ALT200M at different data scales. 
Note that the larger dataset is a superset of the smaller ones.


We use AdamW optimizer with linearly decaying learning rate. During pre-training, the 
batch size is $8192$. The initial learning rate is set to $2\times 10^{-4}$ for the base and large model, and to $1\times10^{-4}$ for the huge model.  The models are trained for $60$ epochs. The maximum length of image regions, tags and caption tokens are $50$, $15$, $20$, respectively. During finetuning, the model is trained for $40$ epochs with batch size $512$. The initial learning rate is $1\times10^{-5}$, $1\times10^{-6}$, and $8\times10^{-7}$ for the base, large, and huge models, respectively. During inference, the caption is generated with beam search and the beam size is $5$. The generation ends when the $\langle \text{EOS} \rangle$ token is predicted, or the maximum length of $20$ tokens is reached. More training details are provided in Appendix.


\subsection{Captioning Results} \label{sec:cap_results}

Results on \nocaps~validation and test sets are shown in~\Cref{tab:nocaps_caption}. By leveraging large-scale pre-training on the automatically collected alt-texts, LEMON has achieved remarkable improvement, especially for out-of-domain images. Compared to the baseline trained on COCO only (row 8), after pre-training on ALT200M (row 12), the CIDEr score is improved by $16.3$ for the in-domain part, and $45.3$ for the out-of-domain part. This evidences that large-scale pre-training improves the model's ability to recognize a wide range of long-tailed visual objects. We also present results of models pre-trained on CC3M and CC12M. Compared to the best reported results on these datasets (row 1, 2), our evaluated CIDEr scores (row 9, 10) are increased by $18.4$ and $13.0$, respectively.
This demonstrates the  performance  improvement in our captioning results brought about by the proposed training scheme when the pre-training dataset is the same.
On the leaderboard\footnote{\url{https://eval.ai/web/challenges/challenge-page/355/leaderboard/1011}} test set, our large and huge models (row 19, 20) both surpassed the top-ranking model (row 18) that is pre-trained on $1.8$B image-text pairs, creating the new state-of-the-art of $114.3$ in CIDEr.
We also achieve the state of the art on other image captioning benchmarks, including COCO Caption and Conceptual Captions, as summarized in Table \ref{tab:coco} and \ref{tab:cc}.

\begin{figure*}[t]
\begin{center}
\includegraphics[trim=5 205 140 0, clip,width=0.92\textwidth]{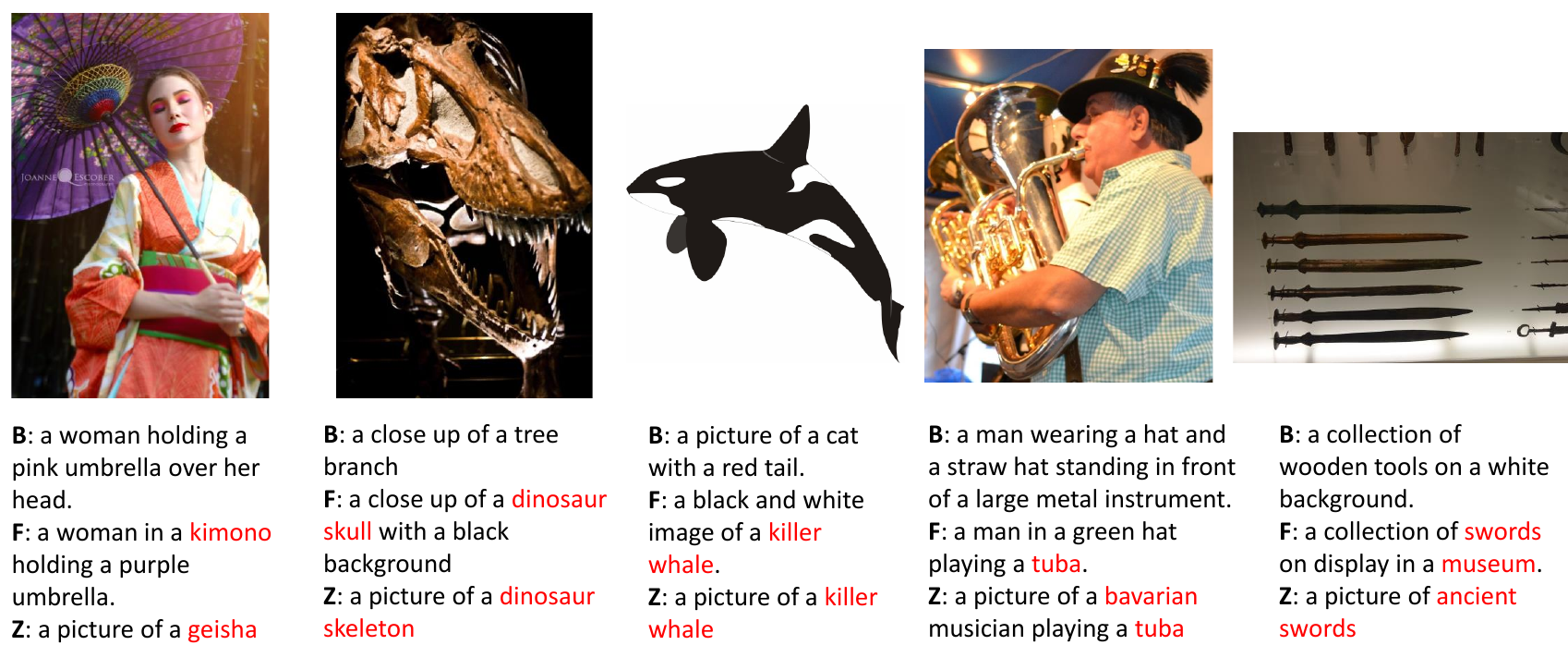}
\vspace{-2mm}
\caption{\textbf{Examples of generated captions on \nocaps~validation set}. \textbf{B}: the baseline model trained on COCO caption only without pre-training. \textbf{F}: the model finetuned on COCO after pre-training on ALT200M. \textbf{Z}: the pre-trained model without finetuning, where we add the prefix ``a picture of'' during inference as the prompt to improve the quality of zero-shot generation following \cite{wang2021simvlm}.}
\label{fig:visual}
\end{center}
\vspace{-5mm}
\end{figure*}

Large-scale pre-training not only benefits VL representation learning, but also equips the model with the capability to zero-shot generalization. We use the pre-trained model to generate captions directly without further finetuning. The prefix ``a picture of'' is added as prompt to improve the quality of generated captions. Some examples are illustrated in \Cref{fig:visual}. The pre-trained model demonstrates strong ability in recognizing various long-tail visual concepts. Compared to the model trained only on small clean set, it shows the knowledge of many fine-grained categories (\emph{e.g.}, ``metal instrument'' vs. ``tuba''), which are learned from the large-scale noisy supervision of alt-texts from web. We also notice that our pre-trained model tends to generate very short descriptions when used in a zero-shot manner, but this is mitigated after finetuning on COCO. We posit that the reason for this is the relatively large proportion of short alt-texts in our pre-training datasets.


\begin{table}    
\small
\centering
\begin{tabular}{lcccccccccc}                                                            \toprule
Model & B@4 & M & C & S \\
\midrule
w/o pre-training~\cite{changpinyo2021conceptual} & - & - & $100.9$ & - \\
pre-trained on CC12M~\cite{changpinyo2021conceptual} & - & - & $105.4$ & - \\
\midrule
LEMON$_{base}$ w/o PT  & $10.1$ & $12.1$ & $104.4$ & $19.0$ \\
LEMON$_{base}$ on CC12M  & $10.1$ & $11.9$ & $108.1$ & $19.8$ \\
\rowcolor{Gray}
LEMON$_{base}$ & $10.1$ & $12.0$ & $111.9$ & $20.5$ \\
\rowcolor{Gray}
LEMON$_{large}$ & $10.8$ & $12.3$ & $117.4$ & $21.0$ \\
\rowcolor{Gray}
LEMON$_{huge}$ & $\bf 13.0$ & $\bf 13.9$ & $\bf 136.8$ & $\bf 23.2$ \\
\bottomrule
\end{tabular}
\vspace{-2mm}
\caption{\textbf{Results on the Conceptual Captions (CC3M) dev set}. All the models are finetuned on CC3M with cross-entropy loss only. We compare with the best results reported on the dev set with and without pre-training. PT: pre-training.}
\vspace{-3mm}
\label{tab:cc}
\end{table}

\begin{figure*}%
    \centering
    \subfloat[\centering pre-training accuracy]{{\includegraphics[trim=15 15 0 15,clip,height=0.2\textheight]{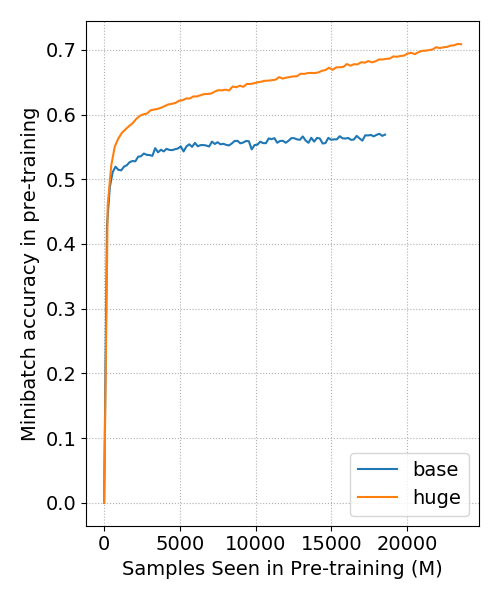} }}%
    \subfloat[\centering finetuned/evaluated on COCO]{{\includegraphics[trim=0 15 14 15,clip,height=0.2\textheight]{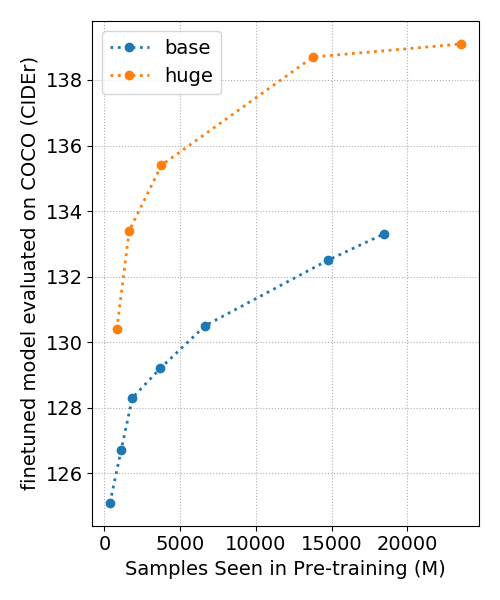} }}%
    \subfloat[\centering finetuned on COCO, evaluated on nocaps]{{\includegraphics[trim=40 10 14 15,clip,height=0.22\textheight]{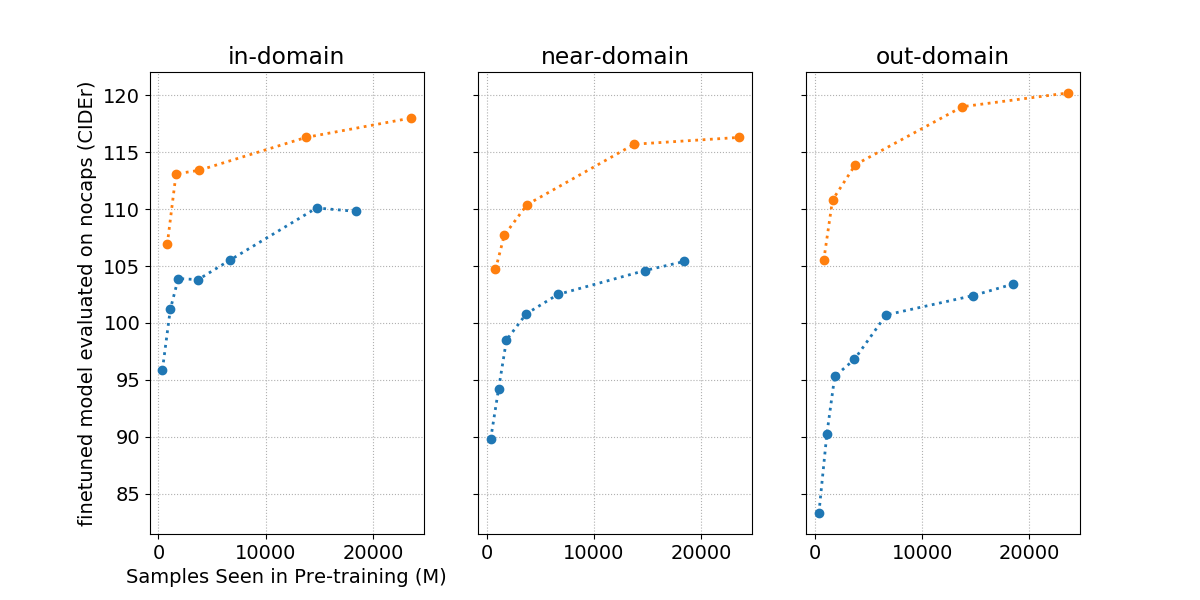} }}%
\vspace{-2mm}
\caption{\textbf{Comparison of sample efficiency for different model sizes}. Figure (a) shows the learning curve in pre-training, measured by the accuracy of cross-entropy loss for masked token prediction. Figures (b) and (c) show the results of finetuned intermediate checkpoints, evaluated on COCO ``Karpathy'' test set and \nocaps~validation set, respectively. The larger model can consistently achieve better results in downstream tasks with far fewer pre-training epochs, especially for out-of-domain data.}%
\label{fig:sample_eff}%
\vspace{-2mm}
\end{figure*}

\subsection{Ablation and Analysis} \label{sec:exp_analysis}

\paragraph{Scaling law: influence of data and model sizes.}
We conduct comprehensive experiments to understand how much gain can be obtained in the downstream tasks by scaling up pre-training. \Cref{fig:data_size} shows the relationship between the number of images used in pre-training and the CIDEr scores evaluated in the downstream captioning tasks. All the models are pre-trained from scratch, and then finetuned on COCO. While all the models can be improved after pre-training with more data, the improvement is obviously less significant for the smaller models than for the larger models. On COCO, the gap between ``small'' and ``large'' models is negligible at $3$M scale, but it becomes 
large
as the data size increases. Moreover, when evaluating on~\nocaps, the gap in out-of-domain set is consistently larger than that in in-domain. This implies the advantage of large models in transferring knowledge from pre-training to downstream tasks, especially when the finetuning data 
are
too limited to cover all test scenarios.

Besides, we observe that the model capacity becomes the performance bottleneck as the amount of available data increases. \Cref{fig:model_size} plots the scaling trend w.r.t. the number of model parameters. When pre-training with $3$M data, the ``base'' size appears to be sufficient, and there is no significant benefit to using larger models. However, with more than $40$M data, the larger models start to outperform the smaller ones by a significant margin. When the data magnitude reaches hundreds of millions, and if the observed trend from ``base'' to ``huge'' can be kept, there is promise in training an even larger model to push the limits of VLP for captioning tasks.

At last, to have a better understanding of the data quality,
we perform pre-training with the same settings on CC12M and the $12$M subset of ALT200M. With the only difference in pre-training data source, the models yield fairly similar results ($0.1$ to $0.3$ differences in CIDEr) on COCO and \nocaps. 
This indicates that our data quality is comparable to that of CC12M.
The observed performance improvement should be attributed to the pre-training scale.

\vspace{-3mm}
\paragraph{Sample efficiency.}
We examine the improvement of learned representations along with the progress of pre-training. Progress is measured quantitatively by the number of image-text paired samples seen in pre-training, \ie, the effective batch size multiplied by the training steps.
In Figure~\ref{fig:sample_eff}, we report the results on COCO Caption after finetuning intermediate pre-trained checkpoints. We also evaluate the finetuned models on~\nocaps, indicating the ability of generalization under domain shift. We present two models in the figure, one with ``base'' size, the other with ``huge'' size. Both models are pre-trained on ALT200M.

We observe that both models continue to improve after seeing more samples in pre-training, but the larger model learns much ``faster''. To achieve similar results in the downstream COCO captioning task, the base model must see more than $2$ to $8$ times more samples in pre-training. This factor is even greater when evaluating on the \nocaps~out-of-domain images. The result of the ``base'' model seeing 19 billion samples is still slightly worse than that of the ``huge'' model seeing 0.8 billion samples. This demonstrates the efficiency of large models in learning from large-scale data, as well as the robustness in generalization.

\begin{table}[t!]
\small
\begin{center}
\begin{tabular}{l@{\hspace{2mm}}|c@{\hspace{2mm}}|c@{\hspace{2mm}}c@{\hspace{2mm}}|c@{\hspace{2mm}}c}
\toprule
\multirow{2}{*}{Arch.} & \multirow{2}{*}{Obj. } & \multicolumn{2}{c|}{COCO} & \multicolumn{2}{c}{CC3M} \\ 
& & CIDEr & SPICE & CIDEr & SPICE \\ \midrule
\multirow{2}{*}{Enc-Dec} & LM &  $\bf 120.9$ & $21.8$ & $94.9$ & $18.1$ \\
        & s2s-MLM &   $120.4$ & $\bf 22.1$ & $99.9$ & $18.9$ \\
\midrule
\multirow{2}{*}{Encoder} & LM &  $119.2$ & $21.5$ & $96.1$ & $18.0$ \\
& s2s-MLM &  $119.9$ & $21.9$ & $\bf 104.4$ & $\bf 19.0$ \\
\bottomrule
\end{tabular}
\end{center}
\vspace{-6mm}
\caption{\textbf{Ablation of models with different architectures, and trained with different objectives}. Results are reported on COCO Caption ``Karpathy'' test split and Conceptual Captions val split. All the models are trained from scratch. s2s-MLM indicates sequence-to-sequence MLM as described in \cref{sec:training_obj}.}
\label{tab:exp_arch}
\vspace{-4mm}
\end{table}

\vspace{-3mm}
\paragraph{Further ablation.}
We compare with other common model structures and training objectives, such as the encoder-decoder transformer model and unidirectional language modeling (LM).
Experiments are conducted with models of ``base'' size as specified in \Cref{tab:arch_details}. For the encoder-decoder structure, we use $6$ encoder layers (with self-attention) followed by $6$ decoder layers (with cross-attention after self-attention), while other model configurations remain unchanged.
The training objectives are illustrated in \Cref{fig:lm}. For each experiment setting, we sweep the hyperparameters, \eg, pre-training epochs from $40$ to $200$, finetuning epochs from $10$ to $60$, and learning rates from $1\times10^{-6}$ to $3\times10^{-5}$. The results of the best hyperparameters are reported.

\begin{figure}[t]
\begin{center}
\includegraphics[trim=0 10 0 10, clip,width=0.4\textwidth]{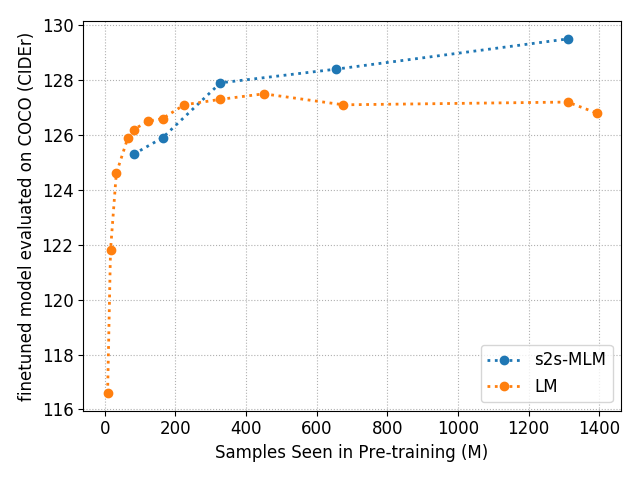}
\vspace{-2mm}
\caption{\textbf{Comparison of different training objectives} by pre-training on CC12M and finetuning on COCO. The models are finetuned from intermediate checkpoints using the same objective as used in pre-training.}
\label{fig:lm_vs_mlm}
\end{center}
\vspace{-6mm}
\end{figure}

We train the models under $4$ different settings on COCO and CC3M, respectively. Results are summarized in \Cref{tab:exp_arch}. On COCO, the differences among the $4$ settings are small ($1.41\%$ relative change in CIDEr), with the worst being $119.2$ from encoder+LM, and the best being $120.9$ from encoder-decoder+LM. In contrast, on CC3M, the difference is much larger ($9.10\%$ relative change in CIDEr). The worst is $94.9$ from encoder-decoder+LM, while the best is $104.4$ from encoder+MLM. As CC3M is collected over the Internet and contains much more noise, we assume that the model that tends to overfit is prone to error when data quality is low, even though it performs well given well-annotated data. 

Moreover, to compare training objectives, we first pre-train the models on CC12M, using s2s-MLM or LM, then finetune the intermediate checkpoints on COCO. As shown in \Cref{fig:lm_vs_mlm}, we observe that although the model trained with LM converges faster at the beginning, it enters saturation early, and does not achieve scores as high as the model using s2s-MLM. We also find that training with LM is very sensitive to learning rates. Given the above results, we choose the s2s-MLM model and the encoder structure to scale up with the noisy pre-training data.


\section{Conclusions}
In this paper, we study the scaling behavior of VLP models for image captioning, and construct our own large-scale dataset ALT200M.
Our experiments show that scaling up pre-training leads to remarkable improvement for the downstream captioning tasks. Our model LEMON has achieved new state-of-the-arts on multiple benchmarks, including COCO Caption,~\nocaps, and Conceptual Captions. LEMON also has impressive capability of recognizing a wide range of long-tail visual objects, even in the zero-shot manner. Moreover, our study on large transformer models indicates that with orders of magnitude larger training data available, the model capacity tends to be the bottleneck. It is a promising direction to train a substantially larger model to take more advantage from the large amounts of alt-text data widely circulated on the Internet. 


{\small
\bibliographystyle{ieee_fullname}
\bibliography{egbib}
}

\appendix
\clearpage
\newpage


\begin{figure*}[t]
\centering
\includegraphics[trim=5 15 5 0, clip,width=0.99\textwidth]{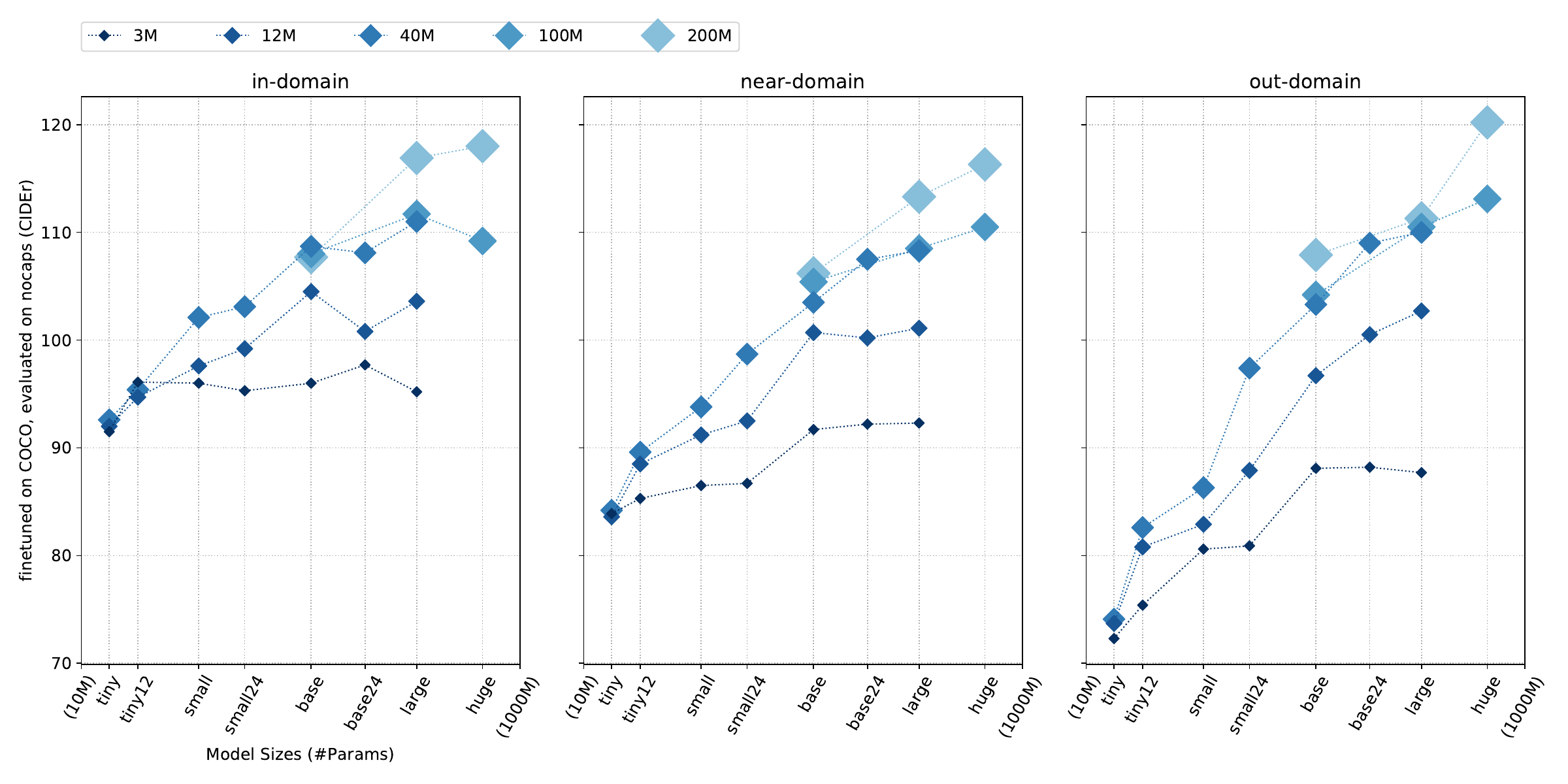}
\caption{\textbf{Image captioning performance on~\nocaps~when upscaling model for each dataset size}. The results of the same models evaluated on COCO are shown in Figure \ref{fig:model_size}. The x-axis plots the number of parameters for each model size (\emph{e.g.}, tiny, small, huge) in a logarithmic scale. The definition of model sizes is detailed in Table \ref{tab:arch_details}. }
\label{fig:model_size_nocaps}
\end{figure*}

\section{Related Work on Image Captioning}
There is a rich literature on image captioning studying different model structures and learning approaches. Recent works have proposed enhanced attention-based models to improve the performance, such as ORT~\cite{herdade2019image}, AoANet~\cite{huang2019attention}, M$^2$ Transformer~\cite{cornia2020meshed}, X-LAN~\cite{pan2020x}, and RSTNet~\cite{zhang2021rstnet}. Besides, researchers have explored to leverage semantic attributes~\cite{yao2017boosting,li2019entangled}, scene graphs~\cite{yang2019auto}, and graph convolutional networks~\cite{yao2018exploring} for captioning.
While those methods focus on learning from well-annotated captions in relatively small datasets, such as COCO Caption, we investigate the impact of data scales with a much larger and noisier dataset. This allows the model to learn more diverse visual concepts, and makes one step further towards in-the-wild captioning.

\section{Comparison of Image-Text Datasets}
Besides CC3M~\cite{sharma2018conceptual} and CC12M~\cite{changpinyo2021conceptual}, we also compare with other web-crawled image-text datasets as described in the following.
\begin{itemize}[leftmargin=*]
\item The dataset used in \textbf{CLIP}~\cite{radford2021learning} has $400$ million image-text pairs. Unlike our dataset crawled from web without re-balancing, their dataset is built upon a set of $500,000$ queries. The queries include all the words occurring at least $100$ times in
the English version of Wikipedia and are augmented with bi-grams. The image-text pairs are searched such that the text includes one of the queries. The final results are also balanced to include up to $20,000$ image-text pairs per query.

\item The dataset used in \textbf{ALIGN}~\cite{jia2021scaling} has $1.8$ billion image-text pairs. A later work SimVLM~\cite{wang2021simvlm} also uses this dataset. The data collection pipeline is similar to that used in Conceptual Captions~\cite{sharma2018conceptual,changpinyo2021conceptual}, but relaxes most cleaning steps. Only some rule-based filters are applied, such as image size, alt-text frequencies, and rare words.

\item The Wikipedia-based Image Text Dataset (\textbf{WIT})~\cite{srinivasan2021wit} is composed of $11.5$ million unique images and $37.6$ million texts. Different from all other listed datasets, it features multilingual texts across $108$ languages. The images and texts are collected from the Wikipedia content pages. It provides texts from multiple sources, such as reference, attribution and
alt-texts, and texts in different languages for the same image.

\item \textbf{WenLan}~\cite{huo2021wenlan} has $30$ million image-text pairs. The web-collected pairs have gone through an elaborate cleaning process. For each data source, they also use topic
models to extract topic words, and analyze the topic distribution to help with selecting desired contents.

\item \textbf{LAION-400M}~\cite{schuhmann2021laion} has $400$ million image-text pairs, and is recently released to public. Instead of applying human designed heuristics in data cleaning, this dataset relies on the CLIP~\cite{radford2021learning} model to filter image-text pairs, where the cosine similarity scores between image and text embeddings are calculated and filtered by $0.3$. As CLIP~\cite{radford2021learning} itself is also trained on noisy image-text pairs, it is yet unknown about the quality of this dataset compared to others cleaned by heuristic-based pipelines.

\end{itemize}

We summarize the characteristics of existing image-text datasets from three perspectives:
\begin{itemize}
    \item \textit{Accessibility}: The datasets CC3M~\cite{sharma2018conceptual}, CC12M~\cite{changpinyo2021conceptual}, WIT~\cite{srinivasan2021wit} and LAION-400M~\cite{schuhmann2021laion} are released to public with the image URL and associated meta files. Other datasets are proprietary.
    \item \textit{Scale}: Except LAION-400M~\cite{schuhmann2021laion}, the other released datasets have tens of millions of images, which are not enough for the scaling purpose.
    \item \textit{Collection pipeline}: LAION-400M~\cite{schuhmann2021laion} is mainly filtered by the CLIP~\cite{radford2021learning} model, while all the other datasets are filtered by a series of rules including expert designed heuristics and/or complicated models. As to the data sources, WIT~\cite{srinivasan2021wit} is collected from Wikipedia. The other datasets are crawled from Internet. 
\end{itemize}

In a nutshell, we construct ALT200M to study the scaling behavior, as no such large-scale datasets are available (LAION-400M~\cite{schuhmann2021laion} is concurrent with ours). Following prior works, we crawl images and the alt attributes from Internet, and apply minimal rule-based filters to retain as many images as possible. Our experiments show that the web-collected data can help to substantially improve the performance of captioning models.

\section{Detailed Hyperparameters}

\begin{table}
\small
\centering
\begin{tabular}{l|cc|cc}
\toprule
\multirow{2}{*}{Model } & \multicolumn{2}{c|}{Pre-training} & \multicolumn{2}{c}{Finetuning}\\
 &  BS & LR & LR ($3$M) & LR ($>3$M) \\
\midrule
tiny & $32768$ & $2\times10^{-3}$  & $2\times10^{-4}$& $2\times10^{-4}$ \\
tiny12 & $32768$ & $1\times10^{-3}$  & $1\times10^{-4}$ & $5\times10^{-5}$\\
small & $16384$ & $5\times10^{-4}$  & $8\times10^{-5}$ & $5\times10^{-5}$\\
small24 & $8192$ & $2\times10^{-4}$  & $5\times10^{-5}$ & $3\times10^{-5}$\\
base & $8192$ & $2\times10^{-4}$  & $3\times10^{-5}$ & $1\times10^{-5}$\\
base24 & $8192$ & $2\times10^{-4}$  & $1\times10^{-5}$ & $5\times10^{-6}$\\
large & $8192$ & $2\times10^{-4}$  & $5\times10^{-6}$ & $1\times10^{-6}$\\
huge & $8192$ & $1\times10^{-4}$  & - & $8\times10^{-7}$\\
\bottomrule
\end{tabular}
\caption{\textbf{Hyperparameters used for all model sizes}. BS: effective batch size. LR: learning rate. LR($3$M) is used for finetuning checkpoints pre-trained on the $3$M subset. LR($>3$M) is used for other finetuning.}
\label{tab:hyperparam}
\end{table}

We include a table of important hyperparameters in Table~\ref{tab:hyperparam} for all model sizes as defined in Table~\ref{tab:arch_details}. All the models are pre-trained for $60$ epochs on subsets of ALT200M, and finetuned for $40$ epochs on COCO with batch size $512$. During pre-training, the learning rate is warmed up for the first $2\%$ steps to the peak value, then linearly decaying to $0$. During finetuning, the learning rate linearly decays from the initial value to $0$ without warm up. We use AdamW optimizer with weight decay $0.05$. The cross-entropy loss is calculated with label smoothing $0.1$.

The evaluation results on COCO ``Karpathy'' test split and~\nocaps~validation set are plotted in Figure~\ref{fig:model_size},~\ref{fig:data_size}, and~\ref{fig:model_size_nocaps}.

\section{More Qualitative Examples}

In this section, we provide more qualitative examples of our web-crawled dataset ALT200M and the captioning results of our LEMON model.

\begin{figure*}[t]
\centering
\includegraphics[trim=0 0 120 0, clip,width=0.9\textwidth]{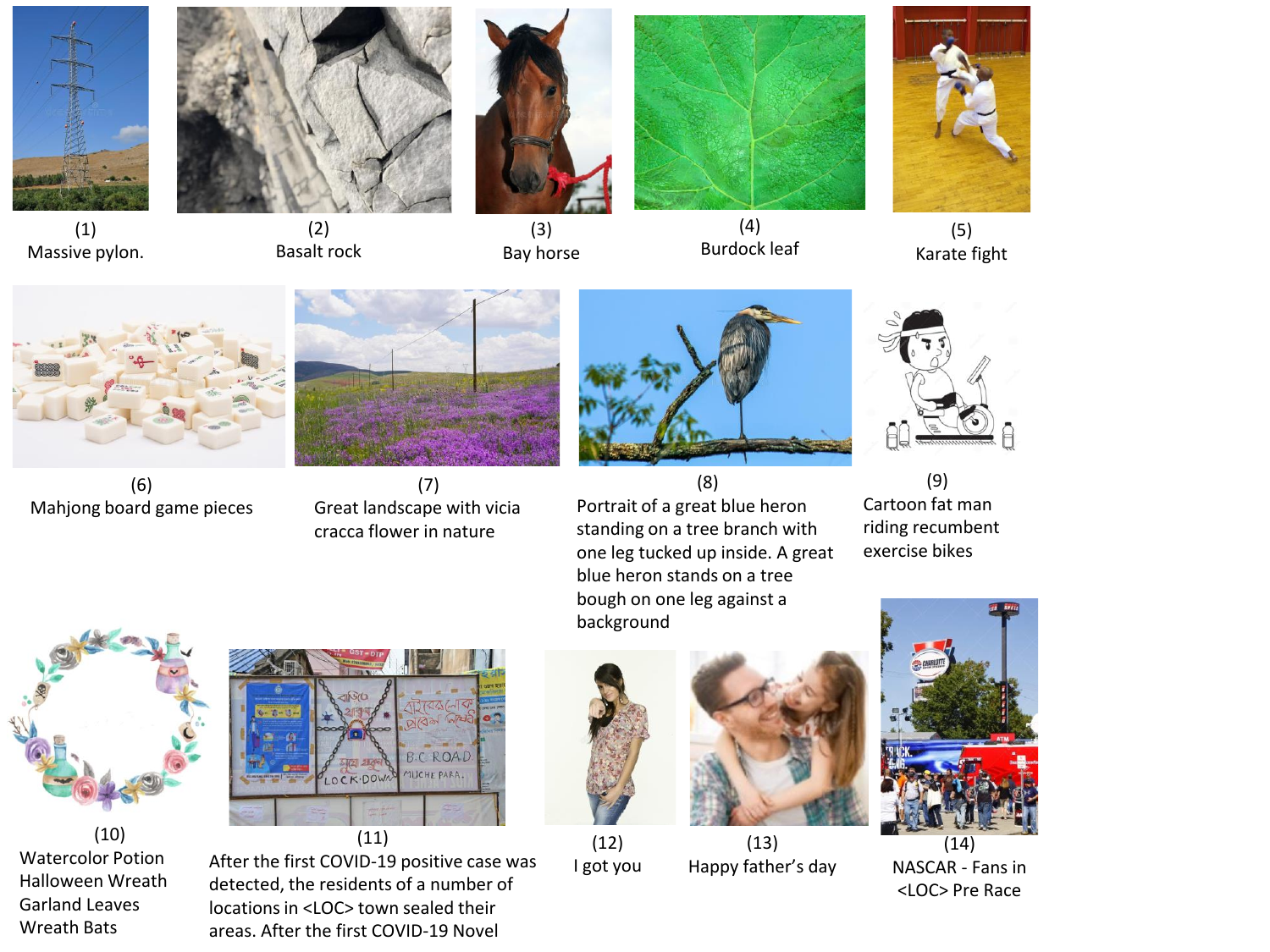}
\caption{\textbf{ALT200M} examples of images and the associated alt attributes. The image-text pairs cover rich visual concepts, but some texts are not well formed (\eg, Figure (10)), or do not directly reflect the image content (\eg, Figure (11) - (14)). }
\label{fig:visual_alt200m}
\vspace{-4mm}
\end{figure*}

Figure~\ref{fig:visual_alt200m} shows some examples of the image-text pairs in ALT200M. While some of the alt attributes are descriptive sentences that can serve as good training targets, \eg, Figure~\ref{fig:visual_alt200m} (7), (8), (9), it is noted that some texts are not semantically well formed, \eg, Figure~\ref{fig:visual_alt200m} (10). Some texts are very short phrases containing only 2 to 4 words, \eg, Figure~\ref{fig:visual_alt200m} (1) - (6). We also observe that some texts do not precisely describe what content is shown in the image, but mention external knowledge or information. For example, Figure~\ref{fig:visual_alt200m} (12) shows a woman pointing at the camera, but the text is ``I got you''. And the text of  Figure~\ref{fig:visual_alt200m} (11) are likely to be extracted from news. These might put challenges for the model to learn from noisy text supervision. However, there are indeed a large variety of (fine-grained) visual objects present in the images and texts, such as burdock leaf, karate, mahjong, and great blue heron. Compared to human-annotated datasets, these web-collected data provide much richer training resources, especially for the long-tailed concepts. 

\begin{figure*}[t]
\centering
\includegraphics[trim=0 130 0 0, clip,width=0.8\textwidth]{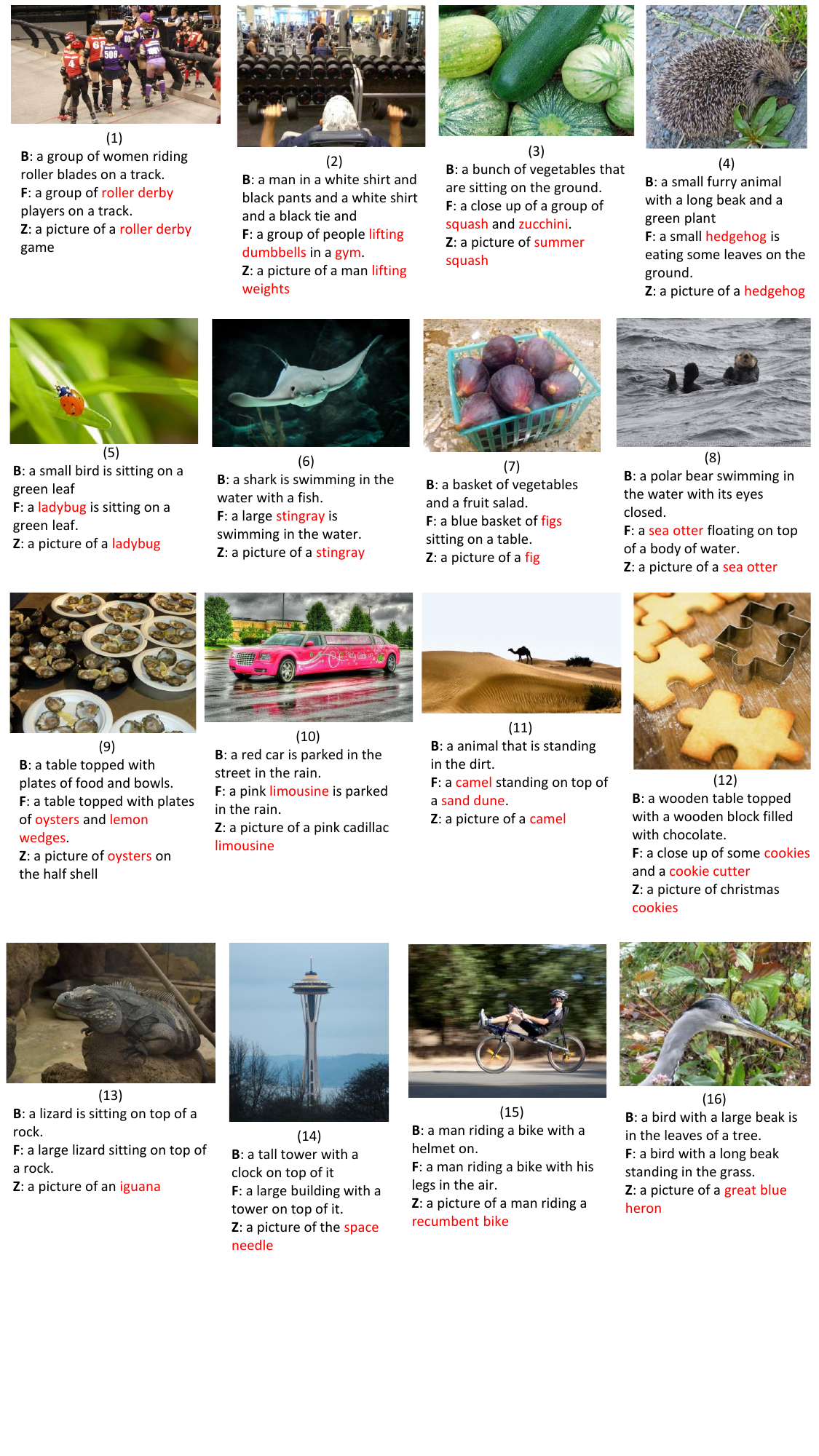}
\caption{More examples of captions generated by our LEMON model in a zero-shot manner (\textbf{Z}) or after finetuning on COCO (\textbf{F}). The notation is the same as described in Figure \ref{fig:visual}.}
\label{fig:visualsupp}
\vspace{-4mm}
\end{figure*}

After training on the ALT200M dataset, our LEMON model has achieved impressive results, even in zero-shot manner. We present more examples of generated captions in Figure~\ref{fig:visualsupp}, in addition to Figure~\ref{fig:visual} in the main paper. Compared to the baseline model trained on COCO only, the LEMON model can recognize much more fine-grained objects, as highlighted in red.

\end{document}